\documentclass[11pt]{article}
\usepackage[utf8]{inputenc}

\usepackage[final]{acl}

\usepackage{times}
\usepackage{latexsym}
\usepackage{booktabs}
\usepackage[T1]{fontenc}

\usepackage{microtype}
\usepackage{amsmath} 
\usepackage{inconsolata}
\usepackage{multirow} 
\usepackage{cleveref}
\usepackage{graphicx}
\usepackage{xspace}
\usepackage{fontawesome}

\usepackage{tikz}
\usetikzlibrary{positioning, fit, shapes.geometric, arrows.meta}
\newcommand{\resource}[1]{\texttt{#1}}
\newcommand{\CLASPAR}{\resource{CLASP-Ar}\xspace}

\title{TTLab at StanceEval-2026: A Cloze-Style Prompting Approach for Arabic-Language Stance Detection (\CLASPAR)}

\author{Bhuvanesh Verma, Ali Abusaleh, Alexander Mehler \\ 
Text Technology Lab (TTLab),  \\
         Goethe University Frankfurt\\
         \{verma,a.abusaleh,mehler\}@em.uni-frankfurt.de
         }

\begin{document}
\maketitle

\begin{abstract}
Arabic-language stance detection remains challenging, and previous shared-task systems have largely relied on multitask learning and ensembles. 
While these systems achieve state-of-the-art performance, their applicability and transferability are limited by the additional complexity introduced by multitask learning.
%
To reduce this complexity, we introduce \CLASPAR, which reformulates the task as cloze-style masked language modeling. 
In this approach, the target, predicted sentiment, and text are combined into a single prompt whose \texttt{[MASK]} prediction is restricted to a verbalizer-constrained label vocabulary.
When evaluated on the StanceEval-2026: Arabic Stance Detection Shared Task, our proposed approach achieves an $F_{\text{avg2}}$ score of $71.36\%$ on Track 1 and $74.14\%$ on Track 2 test sets.
%
The code for \CLASPAR is available at {\href{https://github.com/ENTAILab/ArabicStanceDetection_StanceEval2026}{\faGithub~
TTLab at StanceEval-2026}}
\end{abstract}

\section{Introduction}

Social media platforms have become integral to daily life, facilitating real-time global connectivity and rapid communication, particularly during emergencies \cite{muniz2020social, ghosh2018exploitation}. 
%
While these platforms have broadened participation in public discourse, the rapid growth in user engagement has also accelerated the spread of misinformation, contributing to adverse societal consequences such as increased polarization
\cite{van2021social, kubin2021role}. 
As a result, there is a critical and growing need for automated tools capable of effectively analyzing social media content.

One computational approach to examining online discourse is stance detection, which aims to infer a user's stance toward a specific topic from their textual posts.
%
Typically formulated as a pairwise classification task in which a target topic and its corresponding text are jointly processed, stance detection has attracted considerable research interest \cite{bar2017stance,wei2016pkudblab,lynn2019tweet}.
%
However, most of these efforts have focused primarily on English \cite{zhang2024survey}. 
%
To address this disparity in Arabic, \citet{alturayeif2024stanceeval} introduced the Arabic Stance Detection Shared Task in 2024, using the \resource{Mawqif} dataset \cite{alturayeif-etal-2022-mawqif} to benchmark models on posts from X, formerly Twitter, across three distinct topics.
%
%
StanceEval 2026 ~\cite{Albalawi-etal-2026-stanceeval} builds on this foundation by introducing an extended version of the \resource{Mawqif} dataset. In the previous edition of the task, simple encoder-based architectures achieved highly competitive performance 
%
%
\cite{badran2024alexunlp, hasanaath2024stancecrafters}.


Building on the success of these encoder architectures, we propose a strategy that leverages masked language modeling (MLM) and the bidirectional contextual representations of encoders while incorporating textual sentiment as an explicit feature.
%
More specifically, we propose \CLASPAR, an MLM-based approach that combines principles from pattern-exploiting training (PET) and constrained decoding. It restricts the prediction vocabulary for the masked token to the discrete label space of the stance detection task.
%
The constrained generation strategy of \CLASPAR yields substantial performance improvements over baseline models, achieving an average gain of approximately 2 $F_1$ points across all stance labels.

The present study is organized into the following sections. First, \Cref{sec:background} reviews the theoretical background and describes the task at hand along with the dataset. \Cref{sec:sys_overview} then presents  the underlying methodology and system architecture. The experimental setup and implementation details are covered in \Cref{sec:experiments}. Next, \Cref{sec:result_and_discussion} delivers an in-depth quantitative assessment of the system, further validated via ablation experiments. Lastly, \Cref{sec:conclusion} summarizes the central contributions to bring the paper to a close.

\section{Background}
\label{sec:background}
Target-specific stance detection has been a prominent research area for over a decade. 
Early foundational efforts include the SemEval-2016 shared task for English tweets \cite{mohammad2016semeval} and a subsequent shared task focusing on Chinese microblogs \cite{xu2016overview}. 
Since then, the field has expanded significantly, yielding datasets across diverse languages including Turkish, Czech, Italian, and Arabic \cite{zhang2024survey}.
Following the same line, \citet{Albalawi-etal-2026-stanceeval} introduced StanceEval 2026, a shared task for stance detection in Arabic.

\subsection{Dataset}
This shared task uses \resource{Mawqif-XT} \cite{albalawi2026mawqifxtarabicbenchmarkdataset}, an extended version of \resource{Mawqif} \cite{alturayeif-etal-2022-mawqif}.
It is annotated with stance, sentiment, and sarcasm. 
Each instance in this dataset consists of a target topic under debate and a corresponding controversial text.
%
The controversial text can express support (\textit{favor}) for the target topic, opposition (\textit{against}) to it, or a non-committal (\textit{none}) stance.
%

\subsection{Tracks}
This shared task consists of two tracks.
Track 1 evaluates models on targets observed during training whereas Track 2 evaluates models on unseen targets.
To handle unseen targets under Track 2 and build a robust system for stance detection, we use another dataset.
The second dataset is \resource{ArabicStance-X} \cite{Alkhathlan2025}, which contains 14,477 tweets spanning 17 different topics, including \textit{economy}, \textit{education}, \textit{health}, and \textit{religion}.
%
Structurally, each major topic is divided into subcategories, which contain lists of controversial texts and their corresponding stance labels. The full details of this dataset along with the shared task dataset are shown in \autoref{tab:dataset-stats}.

%
%
The previous edition of Arabic Stance Detection Shared Task in 2024 \cite{alturayeif2024stanceeval} saw systems utilizing multi-task learning to build robust system.
The winning system, \resource{AlexUNLP-BH} \cite{badran2024alexunlp}, used multi-task learning over sentiment and sarcasm, weighted cross-entropy with data augmentation, and an ensemble of models trained with different contrastive loss functions.
The strongest competitors adopted similarly resource-intensive strategies: the MGKM system \cite{alghaslan-almutairy-2024-mgkm} was based on fine-tuned large language models such as \resource{GPT-3.5-Turbo}, whereas \resource{StanceCrafters} \cite{hasanaath2024stancecrafters} combined two \resource{BERT}-based encoders via an attention mechanism in a multitask setup.
These results suggest that auxiliary signals such as sentiment are highly informative for Arabic stance detection.
%
However, top-performing systems obtain these gains at the cost of target-specific models, multiple task heads, or ensembling.
In contrast, our approach incorporates sentiment as a simple input feature in a single cloze-style prompt and replaces the standard classification head with a verbalizer-constrained MLM objective, thereby retaining a single model across all targets and labels.
As a result, our approach largely avoids the complexity introduced by multitask learning and ensembling, making it more suitable for transfer across application scenarios.
\begin{table}[t]

\centering
\small
\begin{tabular}{llr}
\toprule
\textbf{Dataset} & \textbf{Property} & \textbf{Value} \\
\midrule
\multirow{5}{*}{\resource{ArabicStance-X}}
& Usage & Pre-training \\
& Samples & 11,500 \\
& Targets & 17 \\
& Favor & 4,980 \\
& Against/None & 4,452 / 2,068 \\
\midrule
\multirow{6}{*}{{\resource{Mawqif-XT} }}
& Usage & Main task \\
& Train samples & 3,502 \\
& Dev samples & 619 \\
& Train labels & 2148 / 1021 / 333 \\
& Dev labels & 380 / 180 / 59 \\
& Classes & Favor/Against/None \\
\bottomrule
\end{tabular}

\caption{Statistics of the datasets used in this work. \resource{ArabicStance-X}~\cite{Alkhathlan2025} is used only for intermediate stance pre-training, while \resource{Mawqif-XT} \cite{albalawi2026mawqifxtarabicbenchmarkdataset} is used for model development and evaluation.}
\label{tab:dataset-stats}
\end{table}

\section{System Overview}
\label{sec:sys_overview}

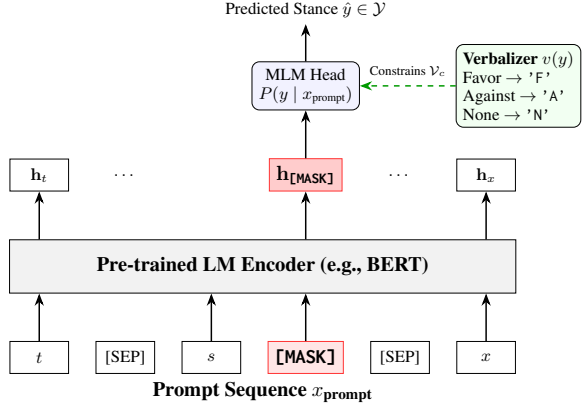
\begin{figure}[t]
\centering
\resizebox{\columnwidth}{!}
{
\begin{tikzpicture}[
    font=\small,
    node distance=0.4cm and 0.5cm,
    block/.style={draw, rectangle, rounded corners, fill=blue!5, minimum height=0.8cm, align=center},
    encoder/.style={draw, rectangle, fill=gray!10, minimum height=1cm, inner sep=0pt},
    token/.style={draw, rectangle, fill=white, minimum height=0.6cm, minimum width=1.1cm},
    mask/.style={token, fill=red!10, draw=red!80, font=\ttfamily\bfseries},
    hstate/.style={token, anchor=south},
    hdots/.style={minimum height=0.6cm, anchor=south},
    arrow/.style={-Stealth, thick}
]
    \node[token] (t_tgt) {$t$};
    \node[token, right=of t_tgt]  (t_sep1) {[SEP]};
    \node[token, right=of t_sep1] (t_snt)  {$s$};
    \node[mask,  right=of t_snt]  (t_mask) {[MASK]};
    \node[token, right=of t_mask] (t_sep2) {[SEP]};
    \node[token, right=of t_sep2] (t_txt)  {$x$};

    \node[fit=(t_tgt)(t_txt), inner sep=0pt,
          label={[font=\bfseries]below:{Prompt Sequence $x_{\text{prompt}}$}}] (inputs) {};

    \node[encoder,
          fit={([yshift=0.9cm]t_tgt.north west) ([yshift=1.9cm]t_txt.north east)},
          label={[font=\bfseries]center:{Pre-trained LM Encoder (e.g., BERT)}}] (bert) {};

    \node[hstate] (h_tgt)  at ([yshift=0.9cm]bert.north -| t_tgt)  {$\mathbf{h}_t$};
    \node[hdots]  (dots1)  at ([yshift=0.9cm]bert.north -| t_sep1) {$\dots$};
    \node[mask, anchor=south, fill=red!20]
                  (h_mask) at ([yshift=0.9cm]bert.north -| t_mask) {$\mathbf{h}_{\text{[MASK]}}$};
    \node[hdots]  (dots2)  at ([yshift=0.9cm]bert.north -| t_sep2) {$\dots$};
    \node[hstate] (h_txt)  at ([yshift=0.9cm]bert.north -| t_txt)  {$\mathbf{h}_x$};

    \node[block, above=0.9cm of h_mask] (mlm) {MLM Head\\$P(y \mid x_{\text{prompt}})$};

    \node[block, fill=green!5, right=1.8cm of mlm, align=left] (verb) {
        \textbf{Verbalizer} $v(y)$ \\
        Favor $\rightarrow$ \texttt{'F'} \\
        Against $\rightarrow$ \texttt{'A'} \\
        None $\rightarrow$ \texttt{'N'}
    };

    \node[above=0.7cm of mlm] (out) {Predicted Stance $\hat{y} \in \mathcal{Y}$};

    \foreach \i in {t_tgt, t_snt, t_mask, t_txt}
        \draw[arrow] (\i.north) -- (\i.north |- bert.south);

    \foreach \i in {h_tgt, h_mask, h_txt}
        \draw[arrow] (bert.north -| \i.south) -- (\i.south);

    \draw[arrow] (h_mask) -- (mlm);
    \draw[arrow, dashed, draw=green!60!black]
        (verb.west) -- node[above=1pt, font=\scriptsize] {Constrains $\mathcal{V}_c$} (mlm.east);
    \draw[arrow] (mlm) -- (out);
\end{tikzpicture}
}
\caption{Overview of \CLASPAR: target $t$, sentiment $s$, and text $x$ form a cloze-style prompt whose \texttt{[MASK]} representation is classified by the MLM head under a verbalizer-constrained softmax over $\mathcal{V}_c = \{\text{`F'}, \text{`A'}, \text{`N'}\}$.}
\label{fig:architecture}
\end{figure}

\CLASPAR's stance detection methodology draws on prompt-based learning paradigms, particularly pattern-exploiting training (PET) \cite{schick2021exploiting}, and constrained decoding techniques \cite{willard2023efficient}. Specifically, we reformulate the standard sequence classification objective as a cloze-style masked language modeling (MLM) task.
%
Given a target $t$, a contextual sentiment label $s$, and an input text $x$, we construct a prompt sequence $x_{\text{prompt}}$ containing a single \texttt{[MASK]} token:
\begin{equation*}
\begin{aligned}
    x_{\text{prompt}} &= \text{Target: } t \text{ Sentiment: } s \\
    &\quad \text{Stance: \texttt{[MASK]} Text: } x
\end{aligned}
\end{equation*}

A BERT-based encoder processes this sequence, and stance is predicted by constraining the model's MLM head to produce outputs only from a predefined subset of vocabulary tokens corresponding to the label space $\mathcal{Y} = {\text{Favor}, \text{Against}, \text{None}}$.
%
Because the natural-language class names may be split into multiple subword tokens by the tokenizer, we define a verbalizer $v: \mathcal{Y} \rightarrow \mathcal{V}$ that maps each class to a single-character token corresponding to its initial letter:
\begin{equation*}
    v(y) = 
    \begin{cases} 
        \text{`F'} & \text{if } y = \text{Favor} \\
        \text{`A'} & \text{if } y = \text{Against} \\
        \text{`N'} & \text{if } y = \text{None} 
    \end{cases}
\end{equation*}
Let $\mathbf{h}_{\texttt{[MASK]}}$ denote the final hidden-state representation of the \texttt{[MASK]} token.
%
The probability assigned to a stance label $y \in \mathcal{Y}$ is then computed by applying a softmax over the constrained vocabulary set $\mathcal{V}_c = \{ \text{`F'}, \text{`A'}, \text{`N'} \}$:
\begin{equation*}
    P(y \mid x_{\text{prompt}}) = \frac{\exp(\mathbf{w}_{v(y)}^\top \mathbf{h}_{\text{[MASK]}})}{\sum_{y' \in \mathcal{Y}} \exp(\mathbf{w}_{v(y')}^\top \mathbf{h}_{\text{[MASK]}})}
\end{equation*}
where $\mathbf{w}_{v(y)}$ represents the weight vector associated with the token $v(y)$ in the pre-trained language model's MLM head. 
The model is optimized end-to-end by minimizing the standard cross-entropy loss between the predicted probabilities and the ground-truth stance label $y^*$:
\begin{equation*}
    \mathcal{L}_{\text{CE}} = -\log P(y^* \mid x_{\text{prompt}})
\end{equation*}

\label{sec:result_and_discussion}
\begin{table}[t]
\centering
\small
\begin{tabular}{lccc}
\toprule
\textbf{Track} & \textbf{Favg2} & \textbf{Favg3} & \textbf{Overall Accuracy} \\
\midrule
Track 1 & 0.7136 & 0.5313 & 0.6761 \\
Track 2 & 0.7414 & 0.6564 & 0.7096 \\
\bottomrule
\end{tabular}
\caption{Performance of \CLASPAR on the final evaluation data of Track 1 (seen targets) and Track 2 (unseen targets).}
\label{tab:track_results}
\end{table}

\section{Experimental Setup}
\label{sec:experiments}
%
To mitigate the class imbalance resulting from the scarcity of the \textit{None} label, we adopt two strategies.
First, we optimize our model using a class-weighted cross-entropy loss function to ensure a balanced penalty distribution during training. 
Secondly, we expand the \resource{Mawqif-XT} dataset. 
Specifically, we generate 1,815 additional \textit{None} instances by randomly pairing controversial texts with mismatched, unrelated targets.

For our base architecture, we utilize the \texttt{aub\-mindlab/bert-base-arabertv02-twitter} encoder \cite{antoun2020arabert} based on the experiments shown in \autoref{ss:encoder_exp}. 
To enhance the model's domain adaptation and task-specific comprehension, we employ a two-stage training paradigm:
\begin{itemize}
    \item \textbf{Phase 1: intermediate pre-training:} The model is initially trained on the \resource{ArabicStance-X} dataset \cite{Alkhathlan2025} for two epochs. 
    Because this auxiliary dataset natively lacks sentiment annotations, we apply \resource{SARF} \cite{Abusaleh:et:al:2026:sarf}, a dedicated Arabic sentiment analysis model, to generate the requisite sentiment values. 
    \item \textbf{Phase 2: fine-tuning:} Subsequently, the encoder is fine-tuned on our augmented \resource{Mawqif-XT}  dataset for up to 20 epochs. 
    To prevent overfitting, we apply early stopping with a patience of five epochs.
\end{itemize}
Across both training phases, we maintain a consistent hyperparameter configuration: a maximum sequence length of 512 tokens, a learning rate of 2e-5, and a batch size of 32.
We evaluate model performance using overall classification accuracy and class-specific F1-scores for the \textit{favor}, \textit{against}, and \textit{none} categories. 
%
%
For our aggregate metrics, we use $F_{\text{avg2}}$ (the macro-average of the two polar classes) as the primary measure, alongside $F_{\text{avg3}}$ to account for all three classes.
%
\section{Results and Discussion}
\autoref{tab:track_results} presents the official evaluation results across both tracks.
Overall, \CLASPAR demonstrates superior performance on Track 2 (unseen targets), achieving an $F_{\text{avg2}}$ of $74.14\%$ and an $F_{\text{avg3}}$ of $65.64\%$.
Conversely, on Track 1 (seen targets), the model yields an $F_{\text{avg2}}$ of $71.36\%$ but experiences a notable drop in $F_{\text{avg3}}$ ($53.13\%$). 
%
%
The substantially lower $F_{\text{avg3}}$ indicates that \CLASPAR struggles to reliably predict the \textit{None} class for this unobserved target.
%
%
%
Overall, we attribute \CLASPAR's strong generalization to Track 2's unseen targets primarily to the pre-training stage, where exposure to a diverse array of target domains enhanced zero-shot target adaptability.
\begin{table}[t]
\centering
\small
\begin{tabular}{lccc}
\toprule
\textbf{Method} & \textbf{$F_{\text{avg2}}$} & \textbf{$F_{\text{avg3}}$} & \textbf{Accuracy} \\
\midrule
Baseline & 85.75 & 71.91 & 83.20 \\

\midrule
\multicolumn{4}{l}{\textit{Without Pre-training}}\\
Base Setup      & 85.95 & 73.62 & 83.94 \\
+ Extended None & \textbf{86.33} & 70.40 & 83.94 \\
+ Class Weights & 85.40 & \textbf{74.23} & 83.36 \\

\midrule
\multicolumn{4}{l}{\textit{With Pre-training}}\\
Base Setup      & 87.35 & 74.87 & \textbf{85.40} \\
+ Extended None & 87.34 & 74.49 & \textbf{85.40} \\
+ Class Weights & 87.26 & \textbf{75.78} & 85.07 \\
\bottomrule
\end{tabular}
\caption{Impact of adding None instances and class weights to \CLASPAR. Full experiment results with multiple seeds can be found in \autoref{app:cls_im}}
\label{tab:class_imbalance}
\end{table}
\subsection{Impact of Class Imbalance Mitigation Strategies}
To systematically evaluate the impact of our class-imbalance mitigation strategies, we conducted multi-seed experiments on the training and development sets.
Notably, the inclusion of augmented None instances (Extended None in \autoref{tab:class_imbalance}) mostly improved performance on the Against label, while consistently harming performance on the None label.
This could mean the augmentation strategy helped create a much clearer boundary between the Against and None labels. 
On the other hand, using class-weighted cross-entropy significantly improved model performance on the None label, evident from the large gains in $F_{\text{none}}$ ($\sim$+7\%) and $F_{\text{avg3}}$ ($\sim$+2\%) over the baseline.
Overall, both approaches boosted performance compared with the baseline and the base configuration, underscoring the effectiveness of these techniques.

\subsection{Impact of Auxilliary Sentiment and Sarcasm Information}
To evaluate the contribution of auxiliary task signals present in the dataset, we conduct an ablation study across three random seeds by incrementally integrating sentiment and sarcasm features. 
As shown in \autoref{tab:aux_exp}, incorporating sentiment information yields the most pronounced performance boost, increasing $F_{\text{avg2}}$ by 1.38 points over the baseline ($86.28\%$ vs. $84.90\%$) and achieving the highest overall accuracy ($84.38\%$). 
While incorporating both auxiliary features yields the highest $F_{\text{avg3}}$ ($72.50\%$), sentiment alone provides the strongest consistent gain across metrics, validating its integration into \CLASPAR.
Detailed class- and target-specific ablation results are provided in \autoref{app:abl_exp}.

\subsection{Sentiment Source}
\label{subsec:sentiment}
Since \resource{ArabicStance-X} nor the official task test data contains no sentiment annotations for data, we generate them with an external Arabic sentiment classifier. 
We compare two sources: a frozen off-the-shelf sentiment-tuned \resource{XLM-RoBERTa} based model~\cite{barbieri-etal-2022-xlm}, and  \resource{SARF}~\cite{Abusaleh:et:al:2026:sarf}, a multi-view Arabic sentiment analyzer  that fuses surface, stemmed, and rooted morphological representations from a shared  \resource{MARBERTv2} encoder via a hybrid CNN--BiLSTM--attention head.
%
%
\resource{SARF} improves the dev  $F_{\mathrm{avg2}}$ from $80.96\%$ to $81.50\%$, so we adopt it as the sentiment source.
\begin{table}[t]
\centering
\small
\begin{tabular}{lccc}
\toprule
\textbf{Ablation Setting} & \textbf{$F_{\text{avg2}}$} & \textbf{$F_{\text{avg3}}$} & \textbf{Accuracy} \\
\midrule
Baseline (None) & 84.90 & 70.83 & 82.61  \\
+ Sentiment     & \textbf{86.28} & 72.36  & \textbf{84.38}  \\
+ Sarcasm       & 84.91 & 71.44 & 82.98  \\
+ Both          & 85.85  & \textbf{72.50} & 83.74  \\
\bottomrule
\end{tabular}
\caption{Ablation analysis on the development set reporting $F_{\text{avg2}}$, $F_{\text{avg3}}$, and Overall Accuracy across 4 settings.}
\label{tab:aux_exp}
\end{table}
\section{Conclusion}
\label{sec:conclusion}
In this work, we present \CLASPAR, a cloze-style masked language modeling framework tailored for stance detection.
Through extensive experiments, we demonstrate that incorporating auxiliary task signals, particularly sentiment features, substantially enhances model performance, improving $F_{\text{avg2}}$ by up to 1.38 points.
Furthermore, our targeted class-imbalance mitigation strategies, such as class weighting, effectively boost performance on the underrepresented \textit{None} class by up to 9 $F_1$ points.
%
Evaluated on the official StanceEval-2026 benchmark, \CLASPAR achieves strong generalization on unseen targets, securing $F_{\text{avg2}}$ scores of $71.36\%$ on Track 1 and $74.14\%$ on Track 2.

\section*{Limitations}
While \CLASPAR avoids the architectural overhead of multitask and ensemble architectures, it exhibits several limitations. 
First, performance on the minority \textit{None} class is notably low and highly volatile, with the standard deviation for $F_{\text{none}}$ exceeding 10 points in certain configurations. 
Second, augmenting \textit{None} instances by pairing texts with unrelated targets represents a naive heuristic that may fail to capture genuine non-committal stances, potentially degrading precision boundaries for the \textit{Against} class. 

\section*{Acknowledgments}
This research is partially funded by the German Research Foundation 
within the Infrastructure Priority Programme \emph{New Data Spaces for the Social Sciences} \href{https://www.dfg.de/de/aktuelles/neuigkeiten-themen/info-wissenschaft/2023/info-wissenschaft-23-20}{(SPP 2431)}, Research-driven Infrastructure for Advanced Survey-related Data (CIRCLET) measure project number ~\href{https://gepris.dfg.de/project/539634240}{539634240} and (Semi-)Automated thematic text classification as a basis for corpus-linguistic value-added services (Project number: \href{https://gepris.dfg.de/project/531750631}{531750631}).

\bibliography{custom}

\begin{thebibliography}{23}
\providecommand{\natexlab}[1]{#1}

\bibitem[{Abusaleh et~al.(2026)Abusaleh, Verma, and Mehler}]{Abusaleh:et:al:2026:sarf}
Ali Abusaleh, Bhuvanesh Verma, and Alexander Mehler. 2026.
\newblock \href {https://doi.org/10.63317/4wj6s3ys5osk} {{TTL}ab at {A}ra{S}ent{E}val: {okSARF} sentiment analysis via root-based fusion for multi-dialectal {A}rabic}.
\newblock In \emph{The 7th Workshop on Open-Source {A}rabic Corpora and Processing Tools ({OSACT}7) with 5 Shared Tasks}, pages 262--268, Palma, Mallorca (Spain). Association for Computational Linguistics.

\bibitem[{Albalawi et~al.(2026{\natexlab{a}})Albalawi, Albadi, Luqman, Ezzini, Kurdi, Yamani, Ashraf, Al-Shaibani, and Alturayeif}]{Albalawi-etal-2026-stanceeval}
Rasha Albalawi, Nuha Albadi, Hamzah Luqman, Saad Ezzini, Maram Kurdi, Asma Yamani, Ahmed Ashraf, Maged Al-Shaibani, and Nora Alturayeif. 2026{\natexlab{a}}.
\newblock Stanceeval-2026: The second stance detection shared task.
\newblock In \emph{Proceedings of the Fourth Arabic Natural Language Processing Conference (ArabicNLP 2026)}, Budapest, Hungary. Association for Computational Linguistics.

\bibitem[{Albalawi et~al.(2026{\natexlab{b}})Albalawi, Albadi, Luqman, Kurdi, Ezzini, Yamani, and Ashraf}]{albalawi2026mawqifxtarabicbenchmarkdataset}
Rasha Albalawi, Nuha Albadi, Hamzah Luqman, Maram Kurdi, Saad Ezzini, Asma Yamani, and Ahmed Ashraf. 2026{\natexlab{b}}.
\newblock \href {https://arxiv.org/abs/2608.09539} {Mawqif-xt: An arabic benchmark dataset for cross-target stance detection}.
\newblock \emph{Preprint}, arXiv:2608.09539.

\bibitem[{Alghaslan and Almutairy(2024)}]{alghaslan-almutairy-2024-mgkm}
Mamoun Alghaslan and Khaled Almutairy. 2024.
\newblock \href {https://doi.org/10.18653/v1/2024.arabicnlp-1.95} {{MGKM} at {S}tance{E}val2024 fine-tuning large language models for {A}rabic stance detection}.
\newblock In \emph{Proceedings of the Second Arabic Natural Language Processing Conference}, pages 816--822, Bangkok, Thailand. Association for Computational Linguistics.

\bibitem[{Alkhathlan et~al.(2025)Alkhathlan, Alahmadi, Kateb, and Al-Khalifa}]{Alkhathlan2025}
Ali Alkhathlan, Faris Alahmadi, Faris Kateb, and Hend Al-Khalifa. 2025.
\newblock Constructing and evaluating {ArabicStanceX}: a social media dataset for arabic stance detection.
\newblock \emph{Front. Artif. Intell.}, 8:1615800.

\bibitem[{Alturayeif et~al.(2024)Alturayeif, Luqman, Alyafeai, and Yamani}]{alturayeif2024stanceeval}
Nora Alturayeif, Hamzah Luqman, Zaid Alyafeai, and Asma Yamani. 2024.
\newblock Stanceeval 2024: The first arabic stance detection shared task.
\newblock In \emph{Proceedings of the Second Arabic Natural Language Processing Conference}, pages 774--782.

\bibitem[{Alturayeif et~al.(2022)Alturayeif, Luqman, and Ahmed}]{alturayeif-etal-2022-mawqif}
Nora~Saleh Alturayeif, Hamzah~Abdullah Luqman, and Moataz Aly~Kamaleldin Ahmed. 2022.
\newblock \href {https://aclanthology.org/2022.wanlp-1.16} {Mawqif: A multi-label {A}rabic dataset for target-specific stance detection}.
\newblock In \emph{Proceedings of the The Seventh Arabic Natural Language Processing Workshop (WANLP)}, pages 174--184, Abu Dhabi, United Arab Emirates (Hybrid). Association for Computational Linguistics.

\bibitem[{Antoun et~al.(2020)Antoun, Baly, and Hajj}]{antoun2020arabert}
Wissam Antoun, Fady Baly, and Hazem Hajj. 2020.
\newblock Arabert: Transformer-based model for arabic language understanding.
\newblock In \emph{LREC 2020 Workshop Language Resources and Evaluation Conference 11--16 May 2020}, page~9.

\bibitem[{Badran et~al.(2024)Badran, Hamdy, Torki, and El-Makky}]{badran2024alexunlp}
Mohamed Badran, Mo’men Hamdy, Marwan Torki, and Nagwa~M El-Makky. 2024.
\newblock Alexunlp-bh at stanceeval2024: Multiple contrastive losses ensemble strategy with multi-task learning for stance detection in arabic.
\newblock In \emph{Proceedings of the Second Arabic Natural Language Processing Conference}, pages 823--827.

\bibitem[{Bar-Haim et~al.(2017)Bar-Haim, Bhattacharya, Dinuzzo, Saha, and Slonim}]{bar2017stance}
Roy Bar-Haim, Indrajit Bhattacharya, Francesco Dinuzzo, Amrita Saha, and Noam Slonim. 2017.
\newblock Stance classification of context-dependent claims.
\newblock In \emph{Proceedings of the 15th Conference of the European Chapter of the Association for Computational Linguistics: Volume 1, Long Papers}, pages 251--261.

\bibitem[{Barbieri et~al.(2022)Barbieri, Espinosa~Anke, and Camacho-Collados}]{barbieri-etal-2022-xlm}
Francesco Barbieri, Luis Espinosa~Anke, and Jose Camacho-Collados. 2022.
\newblock \href {https://aclanthology.org/2022.lrec-1.27} {{XLM}-{T}: Multilingual language models in {T}witter for sentiment analysis and beyond}.
\newblock In \emph{Proceedings of the Thirteenth Language Resources and Evaluation Conference}, pages 258--266, Marseille, France. European Language Resources Association.

\bibitem[{Ghosh et~al.(2018)Ghosh, Ghosh, Ganguly, Chakraborty, Jones, Moens, and Imran}]{ghosh2018exploitation}
Saptarshi Ghosh, Kripabandhu Ghosh, Debasis Ganguly, Tanmoy Chakraborty, Gareth~JF Jones, Marie-Francine Moens, and Muhammad Imran. 2018.
\newblock Exploitation of social media for emergency relief and preparedness: Recent research and trends.
\newblock \emph{Information Systems Frontiers}, 20(5):901--907.

\bibitem[{Hasanaath and Alansari(2024)}]{hasanaath2024stancecrafters}
Ahmed Hasanaath and Aisha Alansari. 2024.
\newblock Stancecrafters at stanceeval2024: Multi-task stance detection using bert ensemble with attention based aggregation.
\newblock In \emph{Proceedings of the Second Arabic Natural Language Processing Conference}, pages 811--815.

\bibitem[{Kubin and Von~Sikorski(2021)}]{kubin2021role}
Emily Kubin and Christian Von~Sikorski. 2021.
\newblock The role of (social) media in political polarization: a systematic review.
\newblock \emph{Annals of the International Communication Association}, 45(3):188--206.

\bibitem[{Lynn et~al.(2019)Lynn, Giorgi, Balasubramanian, and Schwartz}]{lynn2019tweet}
Veronica Lynn, Salvatore Giorgi, Niranjan Balasubramanian, and H~Andrew Schwartz. 2019.
\newblock Tweet classification without the tweet: An empirical examination of user versus document attributes.
\newblock In \emph{Proceedings of the third workshop on natural language processing and computational social science}, pages 18--28.

\bibitem[{Mohammad et~al.(2016)Mohammad, Kiritchenko, Sobhani, Zhu, and Cherry}]{mohammad2016semeval}
Saif Mohammad, Svetlana Kiritchenko, Parinaz Sobhani, Xiaodan Zhu, and Colin Cherry. 2016.
\newblock Semeval-2016 task 6: Detecting stance in tweets.
\newblock In \emph{Proceedings of the 10th international workshop on semantic evaluation (SemEval-2016)}, pages 31--41.

\bibitem[{Muniz-Rodriguez et~al.(2020)Muniz-Rodriguez, Ofori, Bayliss, Schwind, Diallo, Liu, Yin, Chowell, and Fung}]{muniz2020social}
Kamalich Muniz-Rodriguez, Sylvia~K Ofori, Lauren~C Bayliss, Jessica~S Schwind, Kadiatou Diallo, Manyun Liu, Jingjing Yin, Gerardo Chowell, and Isaac Chun-Hai Fung. 2020.
\newblock Social media use in emergency response to natural disasters: a systematic review with a public health perspective.
\newblock \emph{Disaster medicine and public health preparedness}, 14(1):139--149.

\bibitem[{Schick and Sch{\"u}tze(2021)}]{schick2021exploiting}
Timo Schick and Hinrich Sch{\"u}tze. 2021.
\newblock Exploiting cloze-questions for few-shot text classification and natural language inference.
\newblock In \emph{Proceedings of the 16th conference of the European chapter of the association for computational linguistics: main volume}, pages 255--269.

\bibitem[{Van~Bavel et~al.(2021)Van~Bavel, Rathje, Harris, Robertson, and Sternisko}]{van2021social}
Jay~J Van~Bavel, Steve Rathje, Elizabeth Harris, Claire Robertson, and Anni Sternisko. 2021.
\newblock How social media shapes polarization.
\newblock \emph{Trends in cognitive sciences}, 25(11):913--916.

\bibitem[{Wei et~al.(2016)Wei, Zhang, Liu, Chen, and Wang}]{wei2016pkudblab}
Wan Wei, Xiao Zhang, Xuqin Liu, Wei Chen, and Tengjiao Wang. 2016.
\newblock pkudblab at semeval-2016 task 6: A specific convolutional neural network system for effective stance detection.
\newblock In \emph{Proceedings of the 10th international workshop on semantic evaluation (SemEval-2016)}, pages 384--388.

\bibitem[{Willard and Louf(2023)}]{willard2023efficient}
Brandon~T Willard and R{\'e}mi Louf. 2023.
\newblock Efficient guided generation for large language models.
\newblock \emph{arXiv preprint arXiv:2307.09702}.

\bibitem[{Xu et~al.(2016)Xu, Zhou, Wu, Gui, Du, and Xue}]{xu2016overview}
Ruifeng Xu, Yu~Zhou, Dongyin Wu, Lin Gui, Jiachen Du, and Yun Xue. 2016.
\newblock Overview of nlpcc shared task 4: Stance detection in chinese microblogs.
\newblock In \emph{International Conference on Computer Processing of Oriental Languages}, pages 907--916. Springer.

\bibitem[{Zhang et~al.(2024)Zhang, Dai, Niu, Yin, Fan, Wang, Cao, and Huang}]{zhang2024survey}
Bowen Zhang, Genan Dai, Fuqiang Niu, Nan Yin, Xiaomao Fan, Senzhang Wang, Xiaochun Cao, and Hu~Huang. 2024.
\newblock A survey of stance detection on social media: New directions and perspectives.
\newblock \emph{arXiv preprint arXiv:2409.15690}.

\end{thebibliography}

\appendix
\section{Leveraging \resource{ArabicStance-X} for Stance Detection}
\label{ss:encoder_exp}
We investigate two strategies for incorporating \resource{ArabicStance-X} into the training pipeline.
In the \textit{pretrain} setting, the encoder is first fine-tuned on \resource{ArabicStance-X} and subsequently trained on \resource{Mawqif-XT}.
In the \textit{joint} setting, \resource{ArabicStance-X} instances are mixed with the \resource{Mawqif-XT} training data.
Table~\ref{tab:transfer-learning} summarizes the results.

\begin{table}[h]

\centering
\small
\begin{tabular}{lccc}
\toprule
\textbf{Encoder} & \textbf{Plain} & \textbf{Pretrain} & \textbf{Joint} \\
\midrule
\resource{AraBERT} \resource{large-twitter}$^\dagger$ & 84.13 & \textbf{85.24} & 84.42 \\
\resource{MARBERTv2}                       & 82.65 & 82.35 & 82.49 \\
\resource{CAMeLBERT-mix}                   & 79.50 & 80.18 & 81.15 \\
\bottomrule
\end{tabular}
\caption{Impact of transfer learning strategies on development $F_{\mathrm{avg2}}$. 
\textit{Pretrain} denotes intermediate fine-tuning on \resource{ArabicStance-X} followed by training on \resource{Mawqif-XT}. 
\textit{Joint} denotes training on the union of \resource{Mawqif-XT} and \resource{ArabicStance-X}. $^\dagger$ Original shared-task baseline using the large model.}
\label{tab:transfer-learning}

\end{table}

\section{Class Imbalance Mitigation}
\label{app:cls_im}
We perform multi seed experiment to measure the impact of two strategies we adopted for class imbalance. We train model with base configuration, with extended None instances and adding class weights with three seeds (42,123,456). We evaluate these systems on development set (\autoref{tab:stance_results_overall}).  Similarly we also report the impact of these strategies on different targets in \autoref{tab:stance_results_target_specific}.



\begin{table*}[ht]
\centering
\small
\begin{tabular}{lcccccc}
\toprule
\textbf{Method} & \textbf{$F_{\text{favor}}$} & \textbf{$F_{\text{against}}$} & \textbf{$F_{\text{none}}$} & \textbf{$F_{\text{avg2}}$} & \textbf{$F_{\text{avg3}}$} & \textbf{Accuracy} \\ 
\midrule
Baseline & 89.77 & 81.72 & 44.25 & 85.75 & 71.91 & 83.20 \\

\midrule
\multicolumn{7}{l}{\textbf{\CLASPAR}} \\
\midrule
\textit{Without Pre-training} \\
\quad Base Setup & 89.99 ± 0.84 & 81.91 ± 1.03 & 48.96 ± 4.60 & 85.95 ± 0.82 & 73.62 ± 1.78 & 83.94 ± 0.87 \\
\quad + Extended None & 89.94 ± 0.41 & 82.73 ± 0.22 & 38.55 ± 10.81 & 86.33 ± 0.30 & 70.40 ± 3.67 & 83.94 ± 0.42 \\
\quad + Class Weights & 89.66 ± 1.29 & 81.14 ± 0.67 & 51.88 ± 2.13 & 85.40 ± 0.86 & 74.23 ± 0.85 & 83.36 ± 1.33 \\
\addlinespace
\textit{With Pre-training} \\
\quad Base Setup & 90.71 ± 0.26 & 84.00 ± 0.69 & 49.91 ± 5.00 & 87.35 ± 0.46 & 74.87 ± 1.73 & 85.40 ± 0.49 \\
\quad + Extended None & \textbf{90.83 ± 0.22} & 83.85 ± 0.55 & 48.78 ± 6.07 & \textbf{87.34 ± 0.26} & 74.49 ± 1.98 & \textbf{85.40 ± 0.27} \\
\quad + Class Weights & 90.43 ± 0.30 & \textbf{84.09 ± 1.21} & \textbf{52.83 ± 1.75} & 87.26 ± 0.57 & \textbf{75.78 ± 0.93} & 85.07 ± 0.62 \\
\bottomrule
\end{tabular}
\caption{Overall stance detection performance comparison. Our implementation is presented as mean and standard deviation over 5 runs using different seeds. All values are reported as percentages (\%). Baseline uses same encoder as for the \CLASPAR which is AraBERT-large-twitter}
\label{tab:stance_results_overall}
\end{table*}

\begin{table*}[ht]
\centering
\small
\begin{tabular}{lcccccc}
\toprule
\multirow{2}{*}{\textbf{Method}} & \multicolumn{2}{c}{\textbf{Covid Vaccine}} & \multicolumn{2}{c}{\textbf{Digital Trans.}} & \multicolumn{2}{c}{\textbf{Women Emp.}} \\
\cmidrule(lr){2-3} \cmidrule(lr){4-5} \cmidrule(lr){6-7}
& \textbf{$F_{\text{avg2}}$} & \textbf{$F_{\text{avg3}}$} & \textbf{$F_{\text{avg2}}$} & \textbf{$F_{\text{avg3}}$} & \textbf{$F_{\text{avg2}}$} & \textbf{$F_{\text{avg3}}$} \\
\midrule

\midrule
\multicolumn{7}{l}{\textbf{\CLASPAR}} \\
\midrule
\textit{Without Pre-training} \\
\quad Base Setup & 80.78 ± 1.03 & 64.52 ± 2.83 & 86.30 ± 2.89 & 80.99 ± 2.42 & 88.87 ± 1.19 & 70.48 ± 4.44 \\
\quad + Extended None & 82.52 ± 1.00 & 64.63 ± 2.31 & 84.82 ± 2.10 & 74.17 ± 7.77 & 88.15 ± 0.55 & 67.50 ± 4.67 \\
\quad + Class Weights & 78.86 ± 1.71 & 64.38 ± 1.65 & \textbf{87.08 ± 0.92} & \textbf{81.89 ± 1.53} & 88.82 ± 0.99 & 74.89 ± 1.47 \\
\addlinespace
\textit{With Pre-training} \\
\quad Base Setup & 82.30 ± 1.07 & 66.59 ± 3.00 & 86.15 ± 2.75 & 79.92 ± 3.41 & \textbf{90.47 ± 0.72} & 75.15 ± 3.55 \\
\quad + Extended None & \textbf{84.23 ± 1.03} & \textbf{69.44 ± 0.86} & 84.98 ± 1.65 & 77.92 ± 3.31 & 89.19 ± 0.93 & 71.81 ± 3.93 \\
\quad + Class Weights & 82.87 ± 0.96 & 68.39 ± 1.08 & 85.01 ± 1.86 & 79.55 ± 2.30 & 90.46 ± 0.89 & \textbf{77.20 ± 3.00} \\
\bottomrule
\end{tabular}
\caption{Target-specific stance detection performance comparison. Results are shown as mean and standard deviation over 5 runs. All values are reported as percentages (\%).}
\label{tab:stance_results_target_specific}
\end{table*}

\begin{table*}[htbp]
\centering
\small
\resizebox{\textwidth}{!}{%
\begin{tabular}{lcccccc}
\toprule
\textbf{Ablation} & \textbf{$F_{\text{favor}}$} & \textbf{$F_{\text{against}}$} & \textbf{$F_{\text{none}}$} & \textbf{$F_{\text{avg2}}$} & \textbf{$F_{\text{avg3}}$} & \textbf{Accuracy} \\
\midrule
None      & 89.41 $\pm$ 0.30 & 80.39 $\pm$ 2.47 & 42.67 $\pm$ 1.21 & 84.90 $\pm$ 1.38 & 70.83 $\pm$ 0.61 & 82.61 $\pm$ 1.04 \\
Sentiment & \textbf{90.91 $\pm$ 0.29} & \textbf{81.66 $\pm$ 0.69} & 44.51 $\pm$ 4.77 & \textbf{86.28 $\pm$ 0.20} & 72.36 $\pm$ 1.73 & \textbf{84.38 $\pm$ 0.49} \\
Sarcasm   & 89.87 $\pm$ 0.78 & 79.95 $\pm$ 1.48 & 44.50 $\pm$ 4.23 & 84.91 $\pm$ 0.94 & 71.44 $\pm$ 1.57 & 82.98 $\pm$ 0.92 \\
Both      & 90.28 $\pm$ 1.20 & 81.42 $\pm$ 1.39 & \textbf{45.80 $\pm$ 1.03} & 85.85 $\pm$ 1.28 & \textbf{72.50 $\pm$ 1.10} & 83.74 $\pm$ 1.35 \\
\bottomrule
\end{tabular}%
}
\caption{Comprehensive evaluation showing label-specific $F_1$ scores ($F_{\text{favor}}$, $F_{\text{against}}$, $F_{\text{none}}$), macro-averaged scores ($F_{\text{avg2}}$, $F_{\text{avg3}}$), and overall Accuracy across multi-seed experiments on the development set.}
\label{tab:dev_complete_metrics}
\end{table*}

\subsection{Data Augmentation vs. Loss Weighting}
\label{sec:compare_mitigation}
A comparison of data-level (Extended None) and loss-level (Class Weights) approaches shows a clear trade-off between detecting polar stances ($F_{\text{avg2}}$) and overall multi-class performance ($F_{\text{avg3}}$).
As we can see from \autoref{tab:stance_results_overall}, Class Weights consistently improves the minority \textit{none} class, delivering the highest $F_{\text{none}}$ (52.83\% with pre-training) with low cross-seed variance.
In contrast, data augmentation (Extended None) reduces $F_{\text{none}}$ (falling to 38.55\% without pre-training) and leads to pronounced instability ($\sigma = 10.81$).
This indicates that the augmented \textit{none} instances inject feature noise, complicating convergence within the minority class space.

However, Extended None effectively isolates the polar stance classes. 
By artificially saturating the \textit{none} distribution during training, the model establishes a stricter decision boundary for the \textit{favor} and \textit{against} categories, yielding the highest $F_{\text{avg2}}$ (86.33\% without pre-training). 
Class weighting demonstrates the inverse effect, trading fractional $F_{\text{avg2}}$ reductions for global macro-average ($F_{\text{avg3}}$) improvements.

At the target level (\autoref{tab:stance_results_target_specific}), strategy efficacy strongly correlates with the underlying class distributions.
Extended None optimally resolves the \textit{Covid Vaccine} target, which features a balanced polar distribution ($\sim$43\% \textit{favor} and \textit{against}), maximizing both $F_{\text{avg2}}$ (84.23\%) and $F_{\text{avg3}}$ (69.44\%) when pre-trained.
Conversely, for heavily skewed targets where the \textit{none} class is an extreme minority (e.g., $\sim$5\% in \textit{Women Empowerment} and $\sim$10\% in \textit{Digital Transformation}), Class Weights substantially outperforms augmentation by preventing minority class collapse.
Ultimately, class weighting is the superior mechanism for highly imbalanced distributions to maximize global macro-performance, whereas data augmentation is primarily advantageous for isolating stances in targets with balanced polar classes.

\section{Auxiliary Feature Ablations}
\label{app:abl_exp}
The shared task dataset contained additional information like sentiment and sarcasm. We conducted 3 seeds (42, 123, 456) experiments to determine which feature can be helped for Stance detection. \autoref{tab:dev_complete_metrics} shows results on experiments on development set.

\end{document}